\documentclass[conference]{IEEEtran}

\usepackage[T1]{fontenc}
\usepackage{cite}
\usepackage{amsmath,amssymb}
\usepackage{graphicx}
\usepackage{booktabs}
\usepackage{array}
\usepackage{siunitx} % units in the methodology
\usepackage[table]{xcolor}
\usepackage[most]{tcolorbox}
\usepackage{xurl} % lets long URLs in the references break
\usepackage[hidelinks]{hyperref}
\usepackage{orcidlink}

\newtcolorbox{questionbox}[1][]{%
  enhanced, breakable=false,
  colback=black!4, colframe=black!4, boxrule=0pt, arc=2pt,
  borderline west={2pt}{0pt}{black!70},
  left=8pt, right=8pt, top=6pt, bottom=6pt,
  before skip=8pt, after skip=8pt,
  before upper={{\bfseries\footnotesize\MakeUppercase{#1}}\par\smallskip}}

\begin{document}
\bstctlcite{IEEEexample:BSTcontrol}

\title{Mixed Signals: Weight-Space Programming\\for EEG and ECG}

\author{\IEEEauthorblockN{Dara Varam\,\orcidlink{0009-0004-7149-4842},~\textit{Graduate Student Member, IEEE},
Taraneh Aminosharieh Najafi\,\orcidlink{0009-0009-9406-3072}, 
David Atienza\,\orcidlink{0000-0001-9536-4947},~\textit{Fellow, IEEE}}
\IEEEauthorblockA{Embedded Systems Laboratory (ESL), \'Ecole Polytechnique F\'ed\'erale de Lausanne (EPFL), Switzerland\\
Correspondence: dara.varam@epfl.ch \quad Website: \href{https://daravaram.github.io/BioWSP}{daravaram.github.io/BioWSP}}}

\maketitle

\begin{abstract}
Weight-space programming allows language models to read a trained neural network's weights and edit them. This opens the door to ``training'' by reasoning rather than through gradient descent. Prior work has demonstrated its functionality and promise for game-playing networks and fixing specific failures in language models, even extending to diffusion models. In this work, we present its applicability to models trained on biosignal data, specifically EEG and ECG data. We show that frontier models are not only able to debug specific faults planted in biosignal-based models without being told what these faults are, but are even capable of writing all the weights of a network from scratch, given only the architecture, the task, and a minimal set of training subjects. In particular, in one set of experiments, a language model was able to match standard gradient-descent training using $6\times$ fewer training subjects, through reasoning alone and without any gradient descent step. Networks written by Claude Opus 5.5 also scored higher than classic hand-crafted EEG pipelines (alpha band power, and common spatial patterns with linear discriminant analysis) when built from the same 5 subjects. We also highlight some of the limitations, which include the inability of language models to improve already well-trained classification models, and their incapacity to reliably recognize, from the weights alone, the task a network was trained on. Overall, our results indicate that weight-space programming can be useful for small networks, such as those deployed on edge devices and wearables, but the concept still requires further investigation before it can be used in practice.
\end{abstract}

\section{Introduction}

\begin{figure*}[!t]
    \centering
    \includegraphics[width=\textwidth]{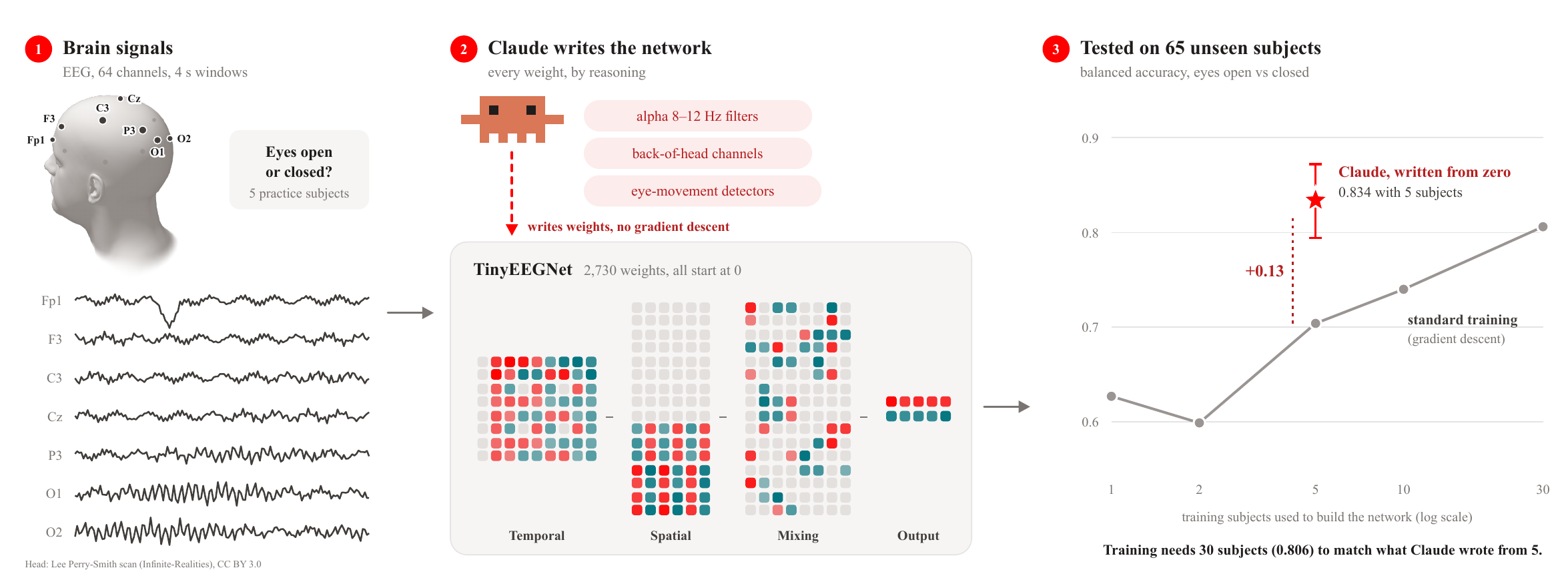}
    \caption{System diagram, showing experiment 5 in this study. A TinyEEGNet network, with all its weights initialized to 0, is written through the reasoning of Claude Opus 5.5 using only 5 training subjects, and matches the same network trained by gradient descent on 30 training subjects.}
    \label{fig:system-diagram}
\end{figure*}

A trained neural network is, ultimately, a long list of numbers that no human chooses. During training, an optimizer sets them through millions of small gradient steps, each bumping the model toward a better fit of labeled examples. This has been the traditional approach to training models. \emph{Weight-space programming} (WSP) proposes a different author for networks' weights: a large language model (LLM). Takehi et al.~\cite{takehi2026sparks} show that such models are able to read the weights as text, reason about what they compute, and write new values, the way a programmer reads and edits code.

The first demonstrations, all very recent at the time of writing, are remarkable. GPT-6 Astra~\cite{openai_gpt6_astra_2026} wrote the weights of game-playing networks for MinAtar~\cite{young2019minatar} from zero, in a single-shot manner. Its Breakout network scored 517.4 points, against 15.8 for a proximal policy optimization (PPO) agent~\cite{schulman2017ppo} trained for 100 million steps~\cite{takehi2026sparks}. The same LLM also taught a reinforcement-learning car to reverse for a run-up, and even raised another language model's mathematical problem solving from 7.9\% to 48.1\% on MATH-500 while changing 0.016\% of its weights. Claude Opus 5.5~\cite{anthropic_claude_2026} edited the text encoders of a diffusion model so that a prompt asking for ``no X'' produces X in 48\% of images instead of 77\%~\cite{chhabria2026negation}. Claude was also able to set every weight of small transformers by hand for ten algorithmic tasks~\cite{singh2026wsp}, and a 27-billion-parameter model running on a laptop named the game a policy network plays from its raw weights in three of four cases~\cite{willison2026qwen}.

Interestingly, the reported failures are as informative as the successes. For example, Takehi et al.~\cite{takehi2026sparks} indicate that no amount of editing was able to improve an already well-trained (``healthy'') Gemma~3~\cite{gemmateam2025gemma3technicalreport}. Critics have also noted that the raw-weight tests used small networks, and that the headline language-model result relied on running the model in Python rather than on reading the weights alone~\cite{goedecke2026superhuman,honnibal2026misleading}.

A pattern is emerging from these first results. WSP works for specific failures with identifiable causes, or for writing short, exact programs. However, there is still no evidence that LLMs can further improve the performance of an already well-trained model. Still, WSP differs from earlier ways of changing weights ``by hand.'' Model-editing methods locate a fact and rewrite it with a closed-form update~\cite{meng2022rome}, and compilers turn a known program into transformer weights~\cite{lindner2023tracr}. With WSP, the LLM does both the analysis and the edit, and decides what to change.

We argue that biosignals are a natural next test, for three reasons: \textbf{(1)} Much of biosignal processing is physics and physiology that we can assume an LLM already knows (through its very large training corpus). A gradient optimizer has to rediscover this from data. \textbf{(2)} Labels are scarce, and subjects vary. EEG and ECG datasets are minuscule compared to image or text datasets, and models degrade on subjects they were not trained on~\cite{roy2019eeg}. Some fail on data from a new site because they learned a shortcut, such as a property of the recording, in place of the physiology~\cite{geirhos2020shortcut,zech2018variable,degrave2021covid}. \textbf{(3)} Most EEG and ECG networks still run offline, after the recording, where their size matters little. However, running them directly on the wearable device is becoming increasingly common, and these networks must then be compact~\cite{lawhern2018eegnet} to fit very constrained hardware~\cite{sopic2018eglass}. This is the scale at which WSP has worked so far, and a scale at which every weight can be read. This leads us to the thesis of this study:

\begin{questionbox}[The open question]
\textbf{Can a language model train a biosignal network by reasoning about physiology, instead of by gradient descent?}\par Given the weights of an EEG or ECG network, and little or no data, can it tell what the network does, make it better, repair it, find out why it fails on unseen data, and write a working network from nothing?
\end{questionbox}

We make a first attempt at answering this question in five experiments, ordered by how much they demand of the LLM (Sec.~\ref{sec:experiments}): 1)~reading a network, 2)~improving it, 3)~repairing a known fault, 4)~debugging an unknown one, and 5)~writing a network from nothing. Throughout, the LLM never takes a gradient step, and every result is judged on data points or recordings it never saw, against the standard training-based fixes a practitioner would apply to the same data.

Our contributions are summarized as follows: 
\begin{enumerate}
    \item We provide the first evaluation of weight-space programming on physiological signals, with five settings across EEG and ECG, from reading a network to writing one, each with success rules fixed in advance using gradient-based baselines, and using three LLMs (one proprietary and two open-weight models).
    \item We show that a network written purely through reasoning matches gradient descent training using $6\times$ fewer training subjects. With no gradient steps and five training subjects, Claude Opus 5.5 wrote all 2,730 weights of a binary EEG classifier (Fig.~\ref{fig:system-diagram}), reaching a balanced accuracy of 0.834 (95\% CI 0.794--0.872), against 0.704 for the same architecture trained traditionally on 5 subjects, and 0.806 when trained on 30. It also scores above classic hand-crafted EEG pipelines built from the same 5 subjects (at most 0.779 [0.729, 0.827]).
    \item We show that language models can act as an effective debugger for biosignal-based models. Without being told the cause, Claude Opus 5.5 exactly named three hidden recording artifacts, and did not invent one in a control case with no artifact. It then switched each shortcut off inside the network. Before the edit, the artifact flipped 75--91\% of the myocardial infarction (MI) predictions, and only 0--2\% after it, which raised MI detection (MI-AUC) from 0.67--0.69 to 0.84--0.86.
    \item We also report the current limits of the approach, as none of the LLMs could recognize the task of a network from its weights alone, improve an already well-trained network, or beat the best gradient-based fix once the cause of a failure is known.
\end{enumerate}

\begin{figure*}[!t]
    \centering
    \includegraphics[width=\textwidth]{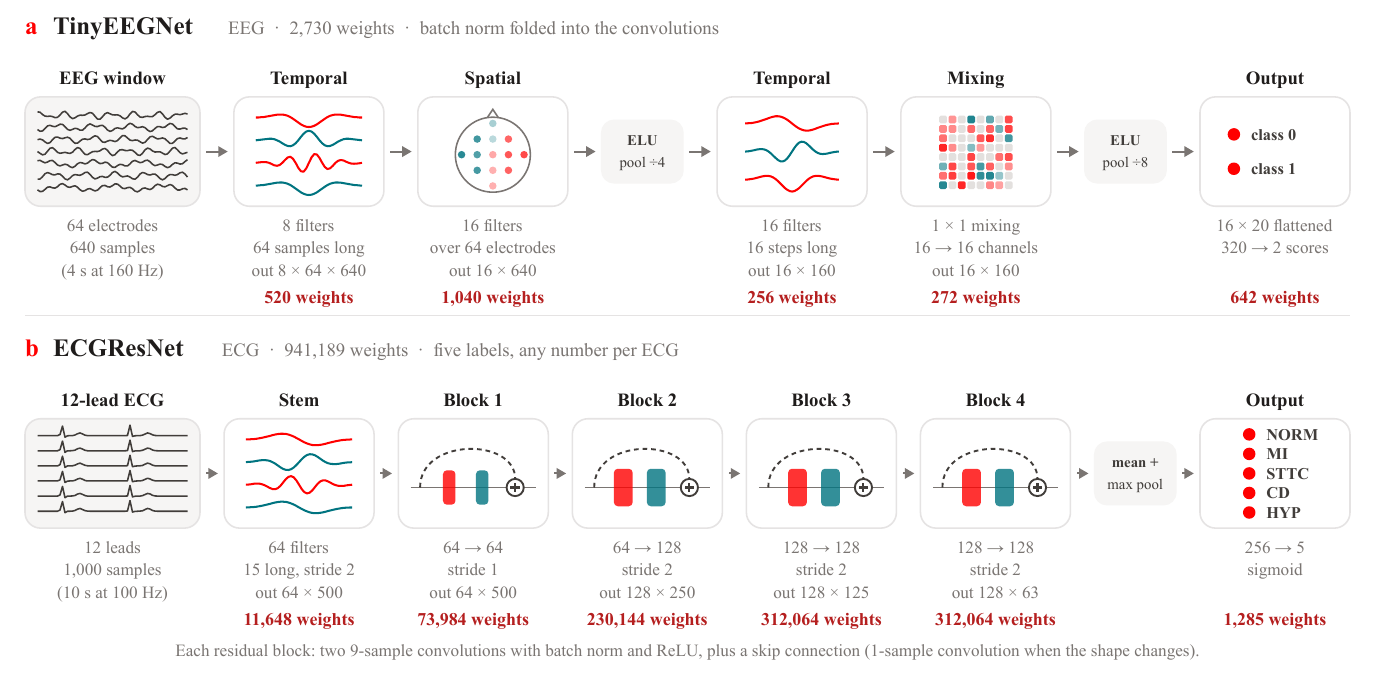}
    \caption{The two architectures, with the output shape and number of weights of each layer. \textbf{(a)}~TinyEEGNet, an EEGNet-style classifier with 2,730 weights, with batch normalization folded into the convolutions. \textbf{(b)}~ECGResNet, a one-dimensional residual network with 941,189 trainable weights (943,877 numbers with the batch-normalization statistics). Filter shapes and weight colors are schematic.}
    \label{fig:networks}
\end{figure*}

\section{Methodology}
\label{sec:setup}

This section describes the data, the networks, the LLMs, the environment and the evaluation protocol used in the five experiments of this study, which are described in Sec.~\ref{sec:experiments}. Throughout, the \emph{LLM} (or \emph{model}) is the language model, and the \emph{network} is the EEG or ECG classifier that it reads, edits or writes. All the prompts and the complete reasoning traces of the LLMs are available on the project website.

\subsection{Data}
\label{sec:data}

\paragraph{EEG} We use the PhysioNet EEG Motor Movement/Imagery dataset~\cite{schalk2004bci2000,goldberger2000physionet}, 64-electrode scalp recordings at \SI{160}{Hz} from 109 subjects. Recordings are band-pass filtered at \SIrange{1}{40}{Hz}, cut into \SI{4}{s} windows and standardized per channel. We define three binary tasks: \textbf{(A)} eyes open against eyes closed at rest, \textbf{(B)} left against right fist movement, and \textbf{(C)} both fists against both feet. We draw 50 subjects and split them into 30 for training, 5 for validation, 5 for calibration, and 10 for testing. The calibration set is the only EEG data the LLM ever sees, called \emph{practice} subjects. No subject appears in two of these sets. For experiment~5 (Sec.~\ref{sec:experiments}), we enlarge the test set to 65 subjects (the remaining 55 eligible subjects plus the original 10).

\paragraph{ECG} We use the PTB-XL dataset~\cite{wagner2020ptbxl}: 21,799 clinical 12-lead ECG recordings of \SI{10}{s} from 18,869 patients, at \SI{100}{Hz}. We keep the 21,388 recordings with a diagnostic label in the five standard superclasses: normal (NORM), myocardial infarction (MI), ST/T change (STTC), conduction disturbance (CD) and hypertrophy (HYP), where a recording can have several labels. We use the recommended split into ten folds, which keeps all recordings of a patient in the same fold: folds 1--8 for training (17,084 recordings), fold 9 as the LLM's practice set (2,146) and fold 10 for testing (2,158). Each lead is standardized with training-set statistics.

\subsection{Networks}
The two networks used in this study across the different experiments are presented in Fig.~\ref{fig:networks}. First, \textbf{TinyEEGNet}, used for all EEG tasks, is a reduced EEGNet~\cite{lawhern2018eegnet} with 2,730 weights: 8 temporal filters (64 samples each) applied to every electrode, 16 spatial filters over the 64 electrodes, a further temporal filter per channel, a $1\times1$ mixing layer, two pooling stages, and a linear layer that scores the two classes. Batch normalization is folded into the preceding convolutions, so every weight is an ordinary multiply-and-add. We train the network with the Adam~\cite{kingma2017adammethodstochasticoptimization} optimizer on the 30 training subjects, with early stopping on the validation set. Training is done for each task with three seeds, giving us a total of 9 baseline networks to compare against, with a balanced accuracy between 0.68 and 0.83. We further train 2 controls with random initialization, and 2 with shuffled labels. 

The second network, \textbf{ECGResNet}, used for all ECG tasks, is a standard one-dimensional residual network with 941,189 trainable weights: a stem convolution with 64 filters across all 12 leads, four residual blocks, average- and max-pooling over time, and a five-output linear layer with sigmoid outputs. We train with AdamW~\cite{loshchilov2019decoupledweightdecayregularization} for 40 epochs on folds 1--8 and keep the epoch with the best macro-AUC on fold~9. We train it with two seeds, which reach a test macro-AUC of 0.917 and 0.918.

\begin{figure*}[!t]
    \centering
    \includegraphics[width=\textwidth]{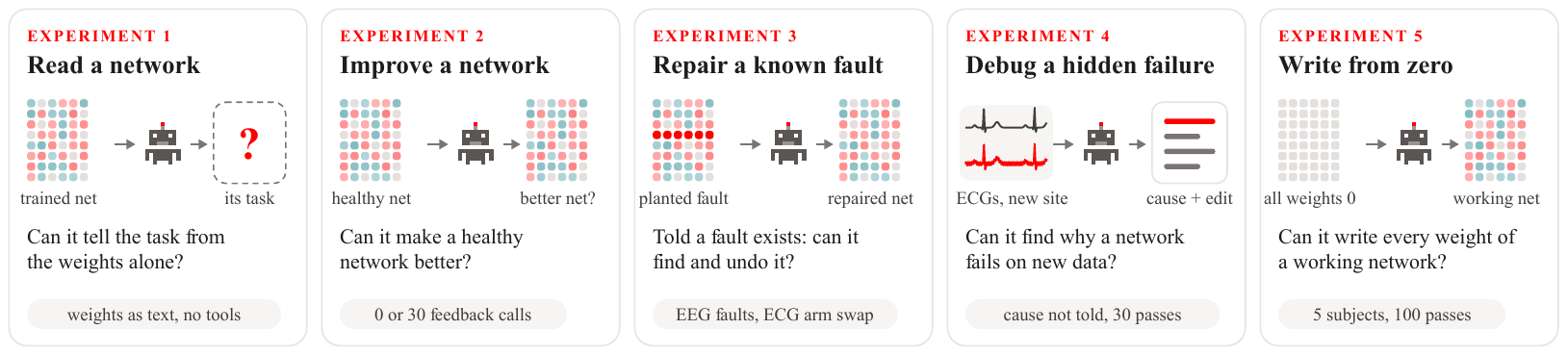}
    \caption{The five experiments, in order of increasing difficulty. In each one, the LLM receives a network (and, depending on the experiment, recordings from practice subjects), works by reasoning only, and hands back an answer or edited weights.}
    \label{fig:experimental-setup}
\end{figure*}

\subsection{Language models}

We compare the performance of a proprietary model, Claude Opus 5.5~\cite{anthropic_claude_2026} (referred to as Opus in the rest of the paper), to two open-weight models, GLM-5.3~\cite{zhipu2026glm} and DeepSeek-V4.1-Flash~\cite{xu2026deepseek}. The same harness is used for all the models, so that all of them see exactly the same prompts and tools. All models are set to the same reasoning effort as Opus in each experiment, and we additionally use gpt-oss-120b~\cite{agarwal2025gpt} as the grader of the textual answers of the models (see Sec.~\ref{sec:evaluation}).

\subsection{Environment and rules}
\label{sec:harness}

\paragraph{Prompt} Each prompt states the task, the format of the data (and necessary pre-processing steps taken), the network's exact computation, and how the result will be measured. We also specify the tools available, and the rules. Importantly, we deliberately leave out the expected answer from the prompt, so that the models never know the outcome. As mentioned, all prompts are identical for each model (see the project website for each prompt).

\paragraph{Tools} Except for experiment~1, where the model can only read and reply, we set up a persistent Python session that holds every weight tensor and can load and save them. The model gets feedback in two formats. 
\begin{enumerate}
    \item An \texttt{evaluate} call (experiments~2 and~3, EEG) returns one number: the balanced accuracy of the current weights on a fresh random batch of 128 trials from the practice subjects, which varies by about $\pm 4$ points between batches.
    \item A \emph{pass} (experiments~3 (ECG), 4 and~5) runs the network on practice data and returns scores, per-class and per-recording (in experiment 5, per-subject) results, internal activations and, on request, a paired comparison between the original and edited network with standard errors. A \emph{probe pass} runs it on signals the model writes itself. Each experiment fixes the number of calls in advance.
\end{enumerate}

\paragraph{Rules} The model is explicitly instructed not to use gradient descent, optimizers of any sort, or any iterative fitting. The network can only be run through \texttt{evaluate} calls or passes. In the ECG experiments, as in Chhabria~\cite{chhabria2026negation}, one-shot linear algebra such as a least-squares solve or a linear discriminant is allowed. In experiment~5, designing filters and analyzing the practice recordings are allowed, but fitting or searching any part of the network against recordings or labels is not. From experiment~3 (ECG) onwards, every tool call must carry a plain-English explanation of the step. Every run also has a fixed limit on passes, turns, code runs, and tokens per reply. The Python session runs in a Linux Landlock sandbox that cannot read the data folders, model files, or test files. 

\subsection{Experimental setup}
\label{sec:experiments}

Fig.~\ref{fig:experimental-setup} summarizes the experimental setups, and Table~\ref{tab:experiments} gives their technical details.

\begin{table*}[!t]
\centering
\footnotesize
\caption{The experiments at a glance. Runs and output-token budgets are per model, and budgets equal Opus's dollar caps converted to tokens. Success rules were fixed before each experiment was run.}
\label{tab:experiments}
\begin{tabular}{@{}>{\raggedright\arraybackslash}p{1.7cm}>{\raggedright\arraybackslash}p{3.3cm}>{\raggedright\arraybackslash}p{2.0cm}>{\raggedright\arraybackslash}p{0.7cm}>{\raggedright\arraybackslash}p{1.4cm}>{\raggedright\arraybackslash}p{6.6cm}@{}}
\toprule
Experiment & The model gets & Feedback & Runs & Budget & Success rule \\
\midrule
1 Read & Trained EEG network as text & None & 52 & 57k & Top-1 $\geq$ 60\% with all weights and names (W1), $\leq$ 30\% without weights and on controls \\
2 Improve & Healthy EEG network & 0 or 30 \texttt{evaluate} & 12 & 42k & Mean gain $\geq$ 2 points, $\geq$ 2 above the placebo, no worse than bias recalibration \\
3 Repair (EEG) & EEG network with one planted fault & 0 or 30 \texttt{evaluate} & 42 & 94k & Recovery $\geq$ 0.5, $\geq$ 0.25 above the placebo, mechanism right in $\geq$ 50\% of runs \\
3 Repair (ECG) & ECG network, arm-swap fault & 30 passes & 1 & 541k & Swapped macro-AUC above a textbook swap check (0.882), correct ECGs unchanged \\
4 Debug & ECG network with a hidden shortcut, or a clean one & 30 passes & 4 & 190k--550k & MI-AUC above the best baseline, other classes no worse \\
5 Write & Empty EEG network & 100 passes & 2 & 250k--461k & Compared with gradient descent on 1--30 training subjects \\
\bottomrule
\end{tabular}
\end{table*}

\paragraph{Experiment 1: reading a network} The models receive a trained TinyEEGNet as text, with every weight rounded to 3 significant figures. No tools are allowed for this experiment. In the first turn, the model describes what signal the network reads, what its task is, and the evidence for its answer. In the second turn, it gives a probability to each of 6 shuffled candidate tasks (3 true tasks, and 3 decoys). To separate reading from prior knowledge, the prompt comes in 5 versions (corresponding to how much information the LLM has about the model), giving us a total of 52 runs per model. 

\paragraph{Experiment 2: improving a network} The models receive a healthy network trained on tasks \textbf{(B)} and \textbf{(C)}, and are asked to raise its balanced accuracy on unseen subjects by editing the weights. Importantly, the model can leave a network unchanged. We test two modes, with feedback (30 \texttt{evaluate} calls) or without it. The model hands back the edited weights and a note on what it changed and why.

\paragraph{Experiment 3: repairing a known fault} The LLMs are told that the network is faulty. In \textbf{TinyEEGNet}, we plant one of four faults in a healthy network after training. Two are visible in the weights (one spatial filter multiplied by 20, called ``hot filter,'' or one output bias shifted up by 3). Two leave the weights looking normal (hence we call them invisible faults): the two class outputs exchanged (\emph{swapped outputs}), or the left and right electrodes exchanged in every spatial filter (\emph{mirrored electrodes}). Each run starts from a fresh copy of the healthy network and contains exactly one of these faults, meaning that faults are never combined. Both feedback and no-feedback options are considered. The LLM is only told that a single fault was introduced into the weights, not where or what it is, and is given the network's validation accuracy before the fault. In \textbf{ECGResNet}, the fault is at the data level: the left- and right-arm electrodes are swapped during recording, a common clinical error~\cite{batchvarov2007lead} that turns
lead I into $-$I, exchanges leads II and III, and exchanges aVR and aVL. Here, the LLM is told exactly what the swap does to the leads.

\paragraph{Experiment 4: debugging failure on unseen data} Here, the models are \textit{not} told the cause of the failure. We train ECGResNet with a planted shortcut: a recording artifact that appears in 98\% of the ECGs with MI, and on 2\% of the others. The network therefore learns that the artifact ``means'' MI. In the unseen data, the artifact
appears on half of the ECGs at random, so the shortcut fails. We use 3 artifacts: mains hum, a calibration pulse at the start of the recording, and a ``muscle filter'' that removes detail above about \SI{20}{Hz}. Networks are considered only if the shortcut clearly hurts performance, and simply removing the artifact is not enough to fix it\footnote{Its MI-AUC is at least 0.05 below a network trained without the shortcut, at least 10\% of its MI calls change, and removing the artifact from the input recovers less than 80\% of the loss.}. A fourth case is used as a control: a clean network tested on a subset of data with patients with more severe conditions. This subset also results in worse performance, but this is not an issue with the recordings. The prompt is the same in all four cases. It says that the network works worse at a second ``hospital,'' above all for MI, and that relying on anything in the recording that does not come from the heart is a defect. The model gets the unseen practice ECGs, labels for 500 of them, and 30 passes. It hands back a diagnosis and edited weights, in one run per case and model.

\paragraph{Experiment 5: writing a network from scratch} The models receive a TinyEEGNet with all 2,730 weights set to zero, and must write every weight so that the network separates the two classes of task \textbf{(A)} (eyes open versus closed) or task \textbf{(B)} (left versus right fist). The prompt gives the exact formula of every layer, the 3D positions of the 64 electrodes, the pre-processing, input means and standard deviations, and how the network will be scored. We also provide access to 5 practice subjects (290 windows for task \textbf{(A)}, 225 for task \textbf{(B)}), up to
100 passes and 400 code runs, and a budget meter shown after every step. Models are expected to hand back a complete network and a description of what each unit does. 

\subsection{Evaluation}
\label{sec:evaluation}

Each experiment has a success rule that is written before running it, and every result is scored by us afterwards with data the language model never saw. We report balanced accuracy for EEG and AUC for ECG. In experiment~1, the runs with no task to find (the prompts without weights and the control networks) are scored against each of the three real tasks and averaged, and a run that gives no answer within its reply limit counts as wrong. In experiment~3, the score is the \emph{recovery}, the share of the accuracy lost to the fault that the repair wins back, where 1 means a full repair.

Every result is compared with networks that are trained with traditional gradient descent. These baselines are the unchanged network after training, a random edit of the same size as the model's edit (a placebo), recalibrating the output biases, fine-tuning on the practice data, and re-training the last layer~\cite{kirichenko2023dfr}. In experiment~4, we add fine-tuning variants that need the diagnosis, such as fine-tuning on ECGs with the artifact removed or randomly added, and we take the best of all baselines on the same 500 labels as the pass mark. In experiment~5, the baseline is the same architecture trained by gradient descent on 1, 2, 5, 10 and 30 training subjects, drawn at random for 1 to 10 (5 seeds), with the original 30 training subjects (3 seeds). We also compare the performance to two hand-crafted EEG pipelines, fitted on the same five practice subjects, namely the relative alpha-band power (8--13 Hz, 8 posterior electrodes) with a single threshold, and common spatial patterns (CSP, 8--30 Hz, 6 filters) followed by a shrinkage linear discriminant analysis (LDA).

\paragraph{Grading} Some outcomes are evaluated as per the model's own plain-English words: the task it names in experiment~1, the fault location and mechanism in experiment~3, and the diagnosis in experiment~4. The gpt-oss-120b model grades the answers of every model against rubrics written before the runs.

\section{Results}

In this section, we go through each of the five experiments of this study. A summary of the main results is included in Table~\ref{tab:overview}. In short, none of the models could improve an already well-trained (healthy) network, but they performed much better whenever there was a clear, concrete fault to find. Planted faults were repaired, hidden recording artifacts were named and switched off, and an empty network was filled with weights that work on subjects the model never saw.

\begin{table}[!htb]
\centering
\footnotesize
\setlength{\tabcolsep}{3pt}
\caption{The results at a glance. Numbers in brackets are the main score of each experiment, explained in its
section below. $^{\dagger}$Number of training subjects that gradient descent needs to do as well, for
eyes open versus closed / left versus right fist.}
\label{tab:overview}
\begin{tabular}{@{}>{\raggedright\arraybackslash}p{3.15cm}ccc@{}}
\toprule
 & Opus & GLM & DeepSeek \\
\midrule
1 \; Read the task & No & No & No \\
2 \; Improve a healthy net & No & No & No \\
3 \; Repair an EEG fault & Yes (0.99) & Partly (0.58) & Yes (0.96) \\
3 \; Repair the ECG swap & Partly (0.870) & No (0.788) & No (0.780) \\
4 \; Name the hidden cause & Yes (3 of 3) & Yes (3 of 3) & Mostly (2 of 3) \\
4 \; Beat best standard fix & No & No & No \\
5 \; Write from zero$^{\dagger}$ & 30 / 10--30 & 5--10 / 2 & 10 / 2--5 \\
5 \; Above hand-crafted$^{\ddagger}$ & Yes / Yes & No / Yes & No / Yes \\
\bottomrule
\end{tabular}
\\[2pt]{\scriptsize $^{\ddagger}$Alpha power and CSP + LDA on the same 5 practice subjects, added after the experiment (Sec.~\ref{sec:evaluation}).}
\end{table}

\subsection{Experiment 1: The models guess, they do not read}

Given a trained network as input, no model could tell what it had been trained to do (Table~\ref{tab:read}). The best score, GLM's 0.33 with electrode names, is no better than its 0.29 on the control networks, which were never trained on any task (leading us to conclude that it was guessing rather than reading the weights). Opus usually recognized the dataset from the electrode layout and the sampling rate, and then picked its best-known task, regardless of what the weights said. In other words, the answer came from what the model already knew about the dataset, not from the weights.

\begin{table}[!htb]
\centering
\footnotesize
\setlength{\tabcolsep}{3pt}
\caption{Experiment 1. Share of runs whose top pick out of six candidate tasks is the right one (chance is 17\%).}
\label{tab:read}
\begin{tabular}{@{}>{\raggedright\arraybackslash}p{3.3cm}cccc@{}}
\toprule
What the model saw & Runs & Opus & GLM & DeepSeek \\
\midrule
A description only, no weights & 8 & 0\% & 4\% & 0\% \\
Weights, no electrode names & 18 & 0\% & 22\% & 11\% \\
Weights and electrode names & 18 & 22\% & 33\% & 0\% \\
Control networks (random or shuffled labels) & 8 & 0\% & 29\% & 12\% \\
\midrule
No answer within the reply limit & 52 & 0 & 5 & 12 \\
\bottomrule
\end{tabular}
\end{table}

\subsection{Experiment 2: A healthy network has little left to give}

\begin{figure}[!htb]
    \centering
    \includegraphics[width=\linewidth]{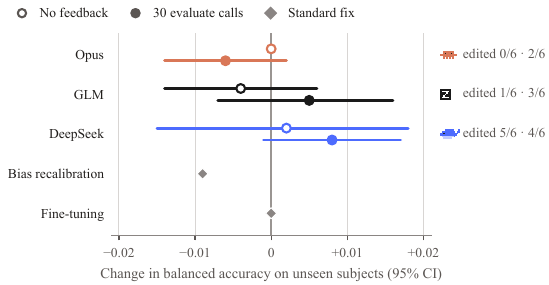}
    \caption{Experiment 2. Change in balanced accuracy on unseen subjects after each model's edit, with no feedback (open marks) and with 30 \texttt{evaluate} calls (filled marks). On the right, how many of the six networks the model chose to edit. Gray diamonds are the standard training-based fixes on the same networks.}
    \label{fig:res-improve}
\end{figure}

None of the models improved a healthy network (Fig.~\ref{fig:res-improve}). With no feedback, Opus left all six networks exactly as they were. This is not a failure of the language models alone. The standard fixes did no better on the same networks, with bias recalibration at $-0.009$ and fine-tuning at $+0.000$. These networks had simply already learned what the data could teach them, which matches what Takehi et al.~\cite{takehi2026sparks} saw with Gemma~3.

% ----------------------------------------------------------------------------

\subsection{Experiment 3: Visible faults are understood, invisible ones are worked around}

\begin{figure}[!htb]
    \centering
    \includegraphics[width=\linewidth]{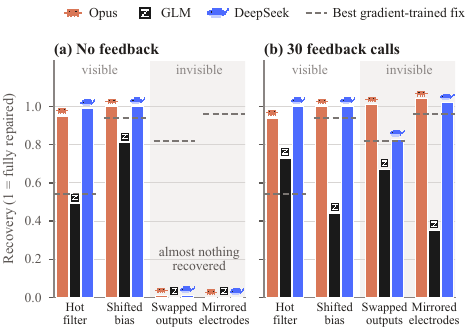}
    \caption{Experiment 3, EEG part. Share of the accuracy lost to each planted fault that the model won back (1 means fully repaired). The dashed line is the best gradient-trained fix for that fault, using the same practice data.}
    \label{fig:res-repair}
\end{figure}

Once the language models were told that something was broken, their outputs differed clearly from the case where they were just told to improve the performance of a healthy network (see Fig.~\ref{fig:res-repair}). The two visible faults were found and undone by Opus and DeepSeek even without any feedback, doing far better than the best gradient-trained fix (0.94--1.00 against 0.54). GLM only partly repaired them (0.49 and 0.81), staying below the baseline performance. Opus and DeepSeek were also able to explain the issues they had found, naming the exact mechanism in 83--100\% of these runs.

On the other hand, the invisible faults were nearly undetectable in the runs with no feedback. None of the LLMs were able to recover the performance, but introducing 30 feedback calls allowed Opus to bring the accuracy back all the way, with DeepSeek close behind. However, they did this without understanding the fault. No model named a swapped output or mirrored electrodes in more than 17\% of runs. Instead, they found edits that the feedback rewarded, and those edits happened to cancel the fault. Averaged over all faults, Opus and DeepSeek recovered 0.99 and 0.96 with feedback, respectively, against 0.40 for the best gradient-trained fix. GLM reached 0.58, and in most of its runs it used up its budget before finishing.

The electrode swap in the ECG was harder (Table~\ref{tab:armswap}). Only Opus made a real difference. Before its edit, the network called 99\% of swapped normal ECGs abnormal, and after it, only 21\%. Its swapped macro-AUC of 0.870 still fell just short of the textbook check, which looks for a mostly negative lead~I and a mostly positive aVR and
swaps the leads back (0.882). The two open models barely moved the network at all.

\begin{table}[!htb]
\centering
\scriptsize
\setlength{\tabcolsep}{2.5pt}
\caption{Experiment 3, ECG part. ECGResNet before and after each model's edit for swapped arm electrodes. To
pass, swapped ECGs must score above the textbook swap check while correct ECGs stay as they were.}
\label{tab:armswap}
\begin{tabular}{@{}>{\raggedright\arraybackslash}p{2.35cm}ccccc@{}}
\toprule
 & Before & Pass if & Opus & GLM & DeepSeek \\
\midrule
Swapped, macro-AUC & 0.766 & $>$ 0.882 & \textbf{0.870} & 0.788 & 0.780 \\
Correct, macro-AUC & 0.9169 & $\geq$ 0.9149 & 0.9167 & 0.9148 & 0.9167 \\
Correct, output changed & 0\% & $\leq$ 2\% & 0.9\% & 6.2\% & 1.0\% \\
Swapped normal, abnormal & 99\% & & \textbf{21\%} & 98\% & 98\% \\
\bottomrule
\end{tabular}
\end{table}

% ----------------------------------------------------------------------------
\subsection{Experiment 4: The models find the hidden cause}

This experiment provides the best outcomes of the whole study. Without being told what the fault was, Opus and GLM named all three artifacts, along with which recordings carry them, how they knew, and where in the network the shortcut lives (Table~\ref{tab:debug}, left). Opus, for example, described the muscle filter as ``a steep 25 Hz low-pass filter (a machine filter setting)'' that is flat up to 21~Hz, at half strength at 25~Hz and empty above 29~Hz, against true values of 20, 25 and 30~Hz. DeepSeek named the mains hum and the calibration pulse, but never a low-pass filter. Importantly, none of the models invented a non-existent cause, either in the three artifact cases or in the control study (where nothing was wrong at all).

\begin{table}[!htb]
\centering
\footnotesize
\setlength{\tabcolsep}{2.4pt}
\caption{Experiment 4. Left: diagnosis points (cause, which recordings carry it, causal evidence, location in the network). Right: share of MI calls that the artifact can still flip, before and after each model's edit.}
\label{tab:debug}
\begin{tabular}{@{}lccccccc@{}}
\toprule
 & \multicolumn{3}{c}{Diagnosis} & \multicolumn{4}{c}{MI calls flipped} \\
\cmidrule(lr){2-4}\cmidrule(l){5-8}
 & Opus & GLM & DS & Before & Opus & GLM & DS \\
\midrule
Mains hum & 4/4 & 4/4 & 3/4 & 81\% & 0\% & 1\% & 0\% \\
Calibration pulse & 4/4 & 4/4 & 4/4 & 91\% & 0\% & 0\% & 6\% \\
Muscle filter & 4/4 & 4/4 & 2/4 & 75\% & 2\% & 3\% & 14\% \\
Control (no artifact) & correct & partly & partly & -- & -- & -- & -- \\
\bottomrule
\end{tabular}
\\[2pt]{\scriptsize DS: DeepSeek-V4.1-Flash.}
\end{table}

\begin{figure}[!htb]
    \centering
    \includegraphics[width=\linewidth]{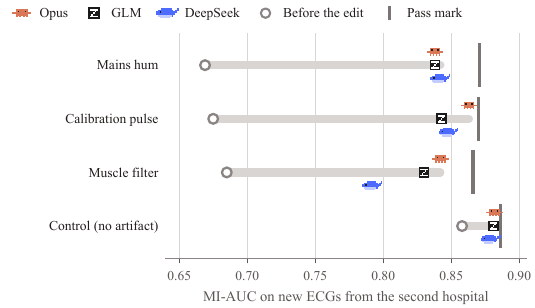}
    \caption{Experiment 4: MI-AUC on new ECGs from the second hospital, before and after each model's edit. The
    gray tick is the pass mark, the best of the standard training-based fixes using the same 500 labeled ECGs.}
    \label{fig:res-debug}
\end{figure}

The models then switched each shortcut off inside the network. Before the edit, the artifact alone flipped 75--91\% of MI calls. After Opus's and GLM's edits it flipped at most 3\%, and after DeepSeek's at most 14\%. MI detection on new ECGs rose from 0.67--0.69 to 0.83--0.86 for every model and every artifact, with the one exception of DeepSeek on the muscle filter (0.79, Fig.~\ref{fig:res-debug}).

However, no edit beat the pass mark set before running the experiment. In the three artifact cases, the best edit fell 0.007--0.029 short of it. Note, though, that those fixes are the ones a practitioner would only think of once the cause is known, such as fine-tuning on ECGs with the artifact added at random, whereas the models found that cause on their own.

\subsection{Experiment 5: A network written from zero}

\begin{figure}[!htb]
    \centering
    \includegraphics[width=\linewidth]{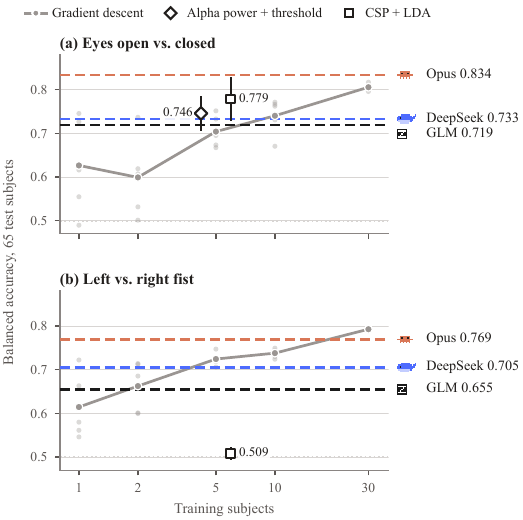}
    \caption{Experiment 5: Balanced accuracy on 65 unseen test subjects of networks written from zero by each model (dashed lines), using recordings from 5 practice subjects, against the same network trained by gradient descent on 1--30 training subjects (gray line, mean over seeds, with each seed as a light dot), and two hand-crafted baselines fitted on the same 5 practice subjects (open markers, with 95\% CIs).}
    \label{fig:res-write}
\end{figure}

For the final and arguably hardest task, we observe, somewhat surprisingly, that every model managed to write a working network from scratch (Fig.~\ref{fig:res-write}). For task \textbf{(A)} (eyes open or closed), Opus's network reached a balanced accuracy of 0.834 [0.794, 0.872] on 65 unseen test subjects. This result is on par with the same network trained with gradient descent on 30 training subjects (0.806, difference $+0.028$ [$-0.016$, $+0.074$]), and well above training on 5 (0.704) or 10 (0.740). For task \textbf{(B)} (left or right fist), it reached 0.769 [0.739, 0.798], above training on 10 subjects but below training on 30. GLM and DeepSeek were also able to write entire networks, but at the level of training on 2 to 10 subjects.

The networks of Opus also score above classic hand-crafted pipelines fitted on the same 5 practice subjects. For eyes open versus closed, relative alpha power with a threshold reached 0.746 [0.708, 0.783] and CSP with LDA reached 0.779 [0.729, 0.827], both below Opus but above the networks of GLM (0.719) and DeepSeek (0.733). For the fist task, CSP with LDA stayed at chance level (0.509 [0.496, 0.522]), as 3 of the 5 practice subjects show little left-versus-right discriminative signal, even individually. Fitted on all 30 training subjects, it reached 0.640, still below every written network.

Given the compact nature of TinyEEGNet, the written networks are particularly easy to interpret. Opus's eyes network looks for alpha rhythm over the back of the head. It filters each electrode into alpha, theta and beta bands, compares neighboring rows of electrodes, and turns each comparison into an amplitude detector. Strong alpha against its neighboring bands means the eyes are closed. The network for fists is built mirror-symmetric, so that swapping the left and right electrodes also flips the decision. It combines three textbook cues: the eyes moving toward the target, the slow potential over the motor cortex on the opposite side, and the drop in beta rhythm there. Interestingly, Opus was not the best judge of its own networks. Opus expected 0.97 for its eyes network, based on the practice recordings it had used to design it (against the 0.834 it actually achieved on unseen subjects). On the fist network, its estimate was 0.774, which is very close to the actual result. 

\paragraph{Cost} Since the two open-weight models, DeepSeek-V4.1-Flash and GLM-5.3, are hosted on our own GPU cluster, writing these networks incurs no API fees for them (only the corresponding GPU time). Claude Opus 5.5, on the other hand, being a proprietary model, cost 8.50 and 9.08~USD in API fees for the two tasks. Importantly, writing a network took 38 to 39 minutes of wall-clock time, while gradient descent trains the same network in 2 to 5 seconds on one NVIDIA A100 GPU (using 5 training subjects). Since the written and the trained networks share the same architecture, their cost at inference time is identical.

\section{Limitations}
We emphasize four main limitations in this study. First, experiments 3 (ECG), 4 and 5 use a single run per model and setting, so their run-to-run variation is not investigated yet. Second, a single LLM (gpt-oss-120b) grades all the textual answers, and a manual check of a subset of these grades, or a panel of several grader models, would strengthen this validation. Third, both PhysioNet datasets are public and likely part of the training corpora of the LLMs (Opus recognized the EEG dataset in experiment 1), so part of what we refer to as prior knowledge of physiology may be knowledge of these specific datasets. An interesting direction would be to repeat the experiments with in-house datasets that are not publicly available. Finally, the two hand-crafted baselines of experiment 5 were added after the LLM runs, so, unlike the other comparisons, their success rule was not fixed in advance. 

\section{Conclusion}
Our results offer mixed signals regarding the applicability of weight-space programming in the biosignal domain. On the one hand, language models are able to write entire networks from scratch for specific applications, outperforming gradient descent with limited training subjects, and, in the case of Opus, even classical hand-crafted baselines. In between, LLMs proved effective at repairing planted faults and at finding hidden recording artifacts, although they did not beat the best gradient-based fix once the cause is known. On the other hand, LLMs struggle to infer the objective of an existing network from its weights alone, and they could not improve an already well-trained network. This indicates that the models succeed not by fitting data, but by recalling structural priors of what a given classifier (e.g., EEG eyes-open versus eyes-closed) should look like. Notably, the open-weight models are not far behind the proprietary one on several of these tasks, despite the differences in general intelligence benchmarks.

This work leaves several directions open, both for biosignal-based models and for weight-space programming as a whole. These findings suggest that WSP could complement, or in some cases bypass, traditional training for tiny models running at the extreme edge in wearables. Furthermore, connecting with work on LLMs as optimizers~\cite{yang2024largelanguagemodelsoptimizers}, an LLM could serve as an initialization pass before gradient descent, streamlining model deployment across biosignals and broader domains.

\section*{Acknowledgment}
This research has been partially supported by the Wyss Center for Bio and Neuro Engineering through the Lighthouse project on Noninvasive Neuromodulation of Subcortical Structures. Artificial Intelligence tools (Claude, Anthropic) were used for figure generation and for writing consistency checks, but all text was written by the authors. The authors assume full responsibility for the content of this paper.

\bibliographystyle{IEEEtran}
\bibliography{references}

\end{document}